\documentclass[letterpaper, 10pt, conference]{ieeeconf}      

\IEEEoverridecommandlockouts                              
\usepackage{graphicx}
\usepackage{url}
\usepackage{setspace}
\usepackage{tikz}
\usetikzlibrary{arrows}
\usetikzlibrary{quotes}
\usetikzlibrary{patterns}
\usetikzlibrary{shapes.geometric}
\tikzset{font={\fontsize{6pt}{12}\selectfont}}
\definecolor{valeocolor}{RGB}{130,230,0}
\tikzstyle{process} = [rectangle, minimum width=1.5cm, minimum height=0.3cm, align=center, draw=black, fill=orange!50]
\tikzstyle{input} = [rectangle, rounded corners, minimum width=1.5cm, minimum height=0.3cm, text centered, draw=black, fill=blue!50]
\tikzstyle{calc} = [ellipse, minimum width=1.0cm, minimum height=0.3cm, text centered, draw=black, fill=valeocolor]
\tikzstyle{arrow} = [thick,->,>=stealth]
\usepackage[lofdepth,lotdepth]{subfig}
\usepackage{pgfplots}
\usepackage{siunitx}
\usepackage{amsmath,amsfonts,amssymb}
\usepackage{slashbox} 
\usepackage{gensymb}
\tikzset{
  every node/.style={
  , execute at begin node=\setlength{\baselineskip}{1em}
  }
}

\title{End to End Vehicle Lateral Control Using a Single Fisheye Camera}

\author{
\authorblockN{Marin Toromanoff\authorrefmark{1}\authorrefmark{2},  Emilie Wirbel\authorrefmark{1}, Fr\'ed\'eric Wilhelm\authorrefmark{1}, Camilo Vejarano\authorrefmark{1}, Xavier Perrotton\authorrefmark{1},  \\Fabien Moutarde\authorrefmark{2}}
\authorblockA{\authorrefmark{1}Valeo Driving Assistance Research France,
\\ name.surname@valeo.com}
\authorblockA{\authorrefmark{2}Center for Robotics, MINES ParisTech, PSL Research University, France
\\name.surname@mines-paristech.fr}
}

\graphicspath{./images}

\begin{document}
\maketitle

\begin{abstract}

Convolutional neural networks are commonly used to control the steering angle for autonomous cars. Most of the time, multiple long range cameras are used to generate lateral failure cases. In this paper we present a novel model to generate this data and label augmentation using only one short range fisheye camera. We present our simulator and how it can be used as a consistent metric for lateral end-to-end control evaluation. Experiments are conducted on a custom dataset corresponding to more than 10000 km and 200 hours of open road driving. Finally we evaluate this model on real world driving scenarios, open road and a custom test track with challenging obstacle avoidance and sharp turns. In our simulator based on real-world videos, the final model was capable of more than 99\% autonomy on urban road. 

\end{abstract}

\section{Introduction}

The ultimate goal for autonomous vehicles is to drive in any environment without any human input. To achieve this, autonomous cars have to analyze their environment using data coming from different sensors and control the car accordingly.
In the most common approach, this task is cut into different modules then fed into a rule-based control algorithm which actually drives the car.
An alternate method would be to make everything in one unique module, i.e. directly taking decisions from raw sensor data without relying on rule-based control. This is the case of end-to-end learning.

In order to train a network to output end-to-end controls, one can train it by trial and error using Reinforcement Learning (RL). Such algorithms are usually trained and tested in a simulator as the work of Perot et al. \cite{Perot2017}. 
Pan et al. \cite{Pan2017} tried to go one step closer to real test by first training a Generative Adversarial Network \cite{Goodfellow} to generate real-looking images from the synthetic images of the simulation and then give these generated images as input to the RL algorithm. But further work has to be done before real car tests could be made safely.

Another way to train an end-to-end algorithm is to use an expert behavior (generally a human driver) as the ground truth. The goal is then to imitate the expert by training a neural network to produce from the raw sensor data the same control output as the expert. This is called imitation learning. The purpose of this work is to use imitation learning to perform lane keeping on open road in diverse locations (with or without road markings, urban streets, country roads, highways) and under diverse weather and lighting conditions. A live demonstration of an end-to-end driving car was also shown at CES 2018 using this work.

The main problem of imitation learning is the
nearly perfect behavior of the expert. Since failure cases (i.e. driving too far away from the desired trajectory) are very rarely encountered, the neural network will never learn to recover from lateral bad positioning of the car, because it has not been provided with enough data.
To tackle this issue, synthetic failure data can be created to train the network to react when those failure cases are encountered. In this article, we will refer to this process as label augmentation. Indeed, the input data and the label are being changed at the same time, contrary to usual data augmentation which changes the input, but not the label.

Bojarski et al. \cite{Bojarski2016d} were the first to successfully train a Convolutional Neural Network (CNN) to infer steering angle from front image by imitation learning, and control the car online (itself inspired from the work done in 1989 by  Pomerleau et al. \cite{Pomerleau1989a}). Their label augmentation is made using three front cameras to simulate small translations and rotations of the car corresponding to lateral failure case. 
Since then, a lot of work has been conducted offline \cite{Xu} \cite{Eraqi2017a} \cite{Chi} on the Udacity challenge and dataset \cite{udacity}.

However there is a huge difference between making an offline prediction and performing an online control. Offline prediction does not take into account the potential accumulation of errors, which means these approaches would not be directly applicable on a real car. Particularly they all use temporal information with a Long Short-Term Memory (LSTM) \cite{Hochreiter1997} and to our knowledge it's still an open problem to handle correctly temporal information with label augmentation.

Furthermore, we can notice that label augmentation was used in all few works performing actual steering control of a real car.
For example Hubschneider et al. \cite{Hubschneider2017b} train a CNN to infer steering angle from a front camera data, complmented with some high level navigation information (turn left or right) to add navigation decision to the lane following paradigm. They generate their label augmentation by shearing the images to simulate a lateral offset. Codevilla et al. \cite{Codevilla2018} use a 3-camera setup for label augmentation, and investigate how to handle specific triggered maneuvers such as turning at an intersection. Yang et al. \cite{Yang2018b} trained a network to infer speed and steering control from front image with the same end-to-end network. It is split between one steering inference branch and another one for speed inference, which share their first layers. They use side cameras to generate lateral failure cases as Bojarski et al. \cite{Bojarski2016d}. We can notice that they are using an LSTM in their network, but only on the speed inference branch. 

Our first contribution is to use only one fisheye camera for both inference and training. This makes it different from \cite{Bojarski2016d} \cite{Yang2018b} who use three cameras, or from Hubschneider et al. \cite{Hubschneider2017b} who shear their images. Our label augmentation model gives us realistic images because we use the camera calibration and intrinsics, and take advantage of the wide field of view of a fisheye camera.
Having only one fisheye camera also reduces integration constraints and makes it identical to the test setup.
Following the idea introduced by Nvidia \cite{Bojarski2016d}, we use this label augmentation to build a simulator based on real-world videos. We make it even more realistic by modeling the physics of the car. This simulator is then extensively used as a validation tool. We argue that our simulator evaluation is more consistent than loss, or the steering angle Mean Square Error (MSE). Indeed, it allows to estimate  the number of unacceptable deviations from the human piloted trajectory (referred to as recoveries in the following).

Our second contribution is an extensive quantitative study, with our realistic simulator, of the influence on the online performance with error accumulation, of label augmentation, failure case generation and training data resampling. To do so, we leverage a dataset containing more than 200 hours of driving, to be compared with 72 hours \cite{Bojarski2016d} and 5 hours \cite{Hubschneider2017b} \cite{Yang2018b}. This bigger dataset makes it possible for our network to have a better generalization capacity, i.e. it can handle new situations better.

Finally we show qualitatively that some uncommon and challenging use cases (sharp turns, object avoidance) can be realized with imitation learning leading to a live demonstration at CES 2018 at Las Vegas.

Our experimental setup is explained in Section~\ref{sec:exp-setup}: target cars and test scenarios, data collection method and network architecture. In Section~\ref{sec:method} our method for training our network is described: first the label augmentation and data selection approach, then the principle of our simulator used to get quantitative test performance. Quantitative evaluations and interpretations are presented in Section~\ref{sec:result}.


\section{Experimental setup}
\label{sec:exp-setup}
In this section, the different cars in which this lateral end-to-end control system has been integrated are presented, along with the target use cases, datasets used, and the neural network architecture with its training hyper-parameters.

\subsection{Target cars}


Three cars are used for data collection and testing. They are equipped with a similar hardware suite, in particular a Controlled Area Network (CAN) bus to obtain the current speed and steering wheel angle, and a fisheye camera. The camera is a Valeo cocoon camera placed in the middle of the bumper. The camera has a 190\degree {}  horizontal field of view, and provides an RGB image with an original resolution of 1280x800 pixels at 30Hz. On each car, the camera is positioned at a similar height (around 50 centimeters).

The two first cars are equipped with an active drive-by-wire control, which makes it possible to actuate the steering wheel, through a MicroAutobox device. One Tegra board of a NVidia DrivePX2 AutoChauffeur is used to run the network inference on both cars. The last one is purely passive, and used to record the initial manual driving database, as described in \ref{sec:10kkm_data} (it is easier for security reasons to use a passive car for human behavior recording).

\subsection{Target scenarios and datasets}

\subsubsection{Open road}
\label{sec:10kkm_data}

\begin{figure}[htpb]
\centering
\subfloat[Urban]{
\includegraphics[width=0.3\columnwidth]{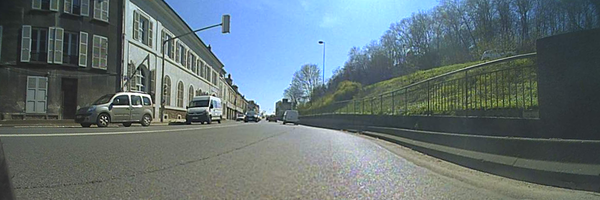}}
\subfloat[Country road]{
\includegraphics[width=0.3\columnwidth]{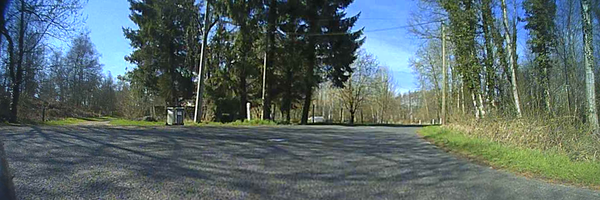}}
\subfloat[Highway]{
\includegraphics[width=0.3\columnwidth]{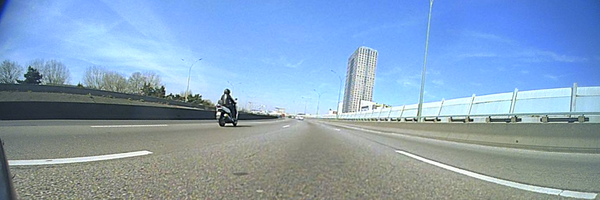}}
\caption{Cylindrical projection of the front fisheye camera, with representative samples of the types of roads traveled}
\label{fig:10kkm_samples}
\end{figure}

To train on openroad data, a dataset has been recorded in the Paris region under numerous weather conditions. It represents more than 10000~km and 200~hours of open road driving.
It contains different kinds of roads: highways, urban streets, country roads etc (see Figure~\ref{fig:10kkm_samples}). This dataset was split into a training set (around 10 million images) and a testing set (around 3 million). All images where the driver wanted to turn or to change lane are removed, based on the car blinkers. Similarly, very low speed images are removed, to avoid specific situations such as parking lots, or crowded urban areas. This is because the target of this work is lane keeping, so these situations need not be handled and could impair the training.

\subsubsection{Test track}
\label{sec:track-data}
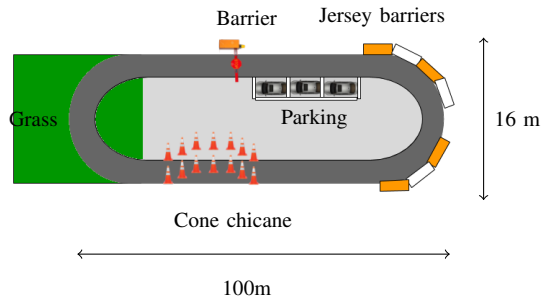
\begin{figure}[h]
   \centering
   \begin{tikzpicture}[scale=0.9, every node/.style={scale=0.9}, font=\small]
   		\node(img) {\includegraphics[width=0.9\columnwidth]{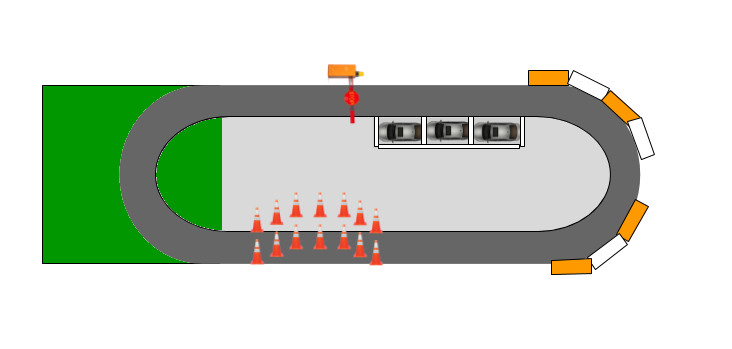}};
        \node(barrier) at (0, 1.5) {Barrier};
        \node(chicane) at (-0.2,-1.5) {Cone chicane};
        \node(grass) at (-3.15,0.0) {Grass};
        \node(parking) at (1.0, 0.0) {Parking};
        \node(jersey) at (2.0,1.5) {Jersey barriers};
        \draw[<->] (-2.5,-2.0) -- (3.0,-2.0);
        \node at (0,-2.5) {100m};
        \draw[<->] (3.5, -1.2) -- (3.5, 1.2);
        \node at (4.0, 0) {16 m};
   \end{tikzpicture}
	\caption{Test track (not to scale) \label{fig:CES_track}}
\end{figure}

As illustrated in Figure~\ref{fig:CES_track}, the test track is composed of two sharp turns (16m in diameter), one dynamic barrier, one straight section and a road deviation through a chicane of traffic cones.
The lines are masked in the two sharp turns to show that we could even handle situations where classic lane based control would have difficulties.

The target here is to adapt the car steering: follow the lane in the straight line, take both sharp turns (in which steering wheel
angle could go up to 500\degree), 
and handle the cones chicane correctly. The longitudinal behavior of the car is controlled manually or by another network (which was the case for the live demonstration).

The described test track contains scenarios that are not common on open road. Therefore some data had to be collected specifically on the test track to learn those unusual situations. Another focus of those recordings is to gather as much variability as possible in terms of lighting and weather conditions, to gain robustness in the final behavior.

We recorded the data as follows: 10 lap series which were reserved for training, and shorter 3 lap series which were kept for validation and testing. The sessions were spaced by roughly one hour over the day, from sunrise to full night. For each test track location, about 5 days of recordings were done, with around 8 recording sessions per day, resulting in roughly 360k training images and 110k testing images.

\subsection{Neural Network  architecture}
Our CNN takes a 200x66 image as input. It contains 10 layers and is close to the one from Nvidia \cite{Bojarski2016d}. It is described in Table~\ref{tab:network-arch}. The hyperparameters for the network are described in Table~\ref{tab:hyperparams} (note that some dropout was added to the fully convolutional layers during training).


\begin{table}[htbp]
\begin{center}
\caption{Network architecture with layers type, parameters and size (Conv - convolutional, FC - fully connected)}
\label{tab:network-arch}
\begin{tabular}{r|ccc}
Name & Size & Filters number & Stride \\
\hline
Normalization & 200x66x3 & N/A & N/A \\
Conv \#1 & 5x5 & 24 & 2 \\
Conv \#2 & 5x5 & 36 & 2 \\
Conv \#3 & 5x5 & 48 & 2 \\
Conv \#4 & 3x3 & 64 & 1 \\
Conv \#5 & 3x3 & 64 & 1 \\
FC \#1 & 1152x1164 & N/A & N/A \\
FC \#2 & 1164x100 & N/A & N/A \\
FC \#3 & 100x50 & N/A & N/A \\
FC \#4 & 50x10 & N/A & N/A \\
FC \#5 & 10x1 & N/A & N/A \\
\end{tabular}
\end{center}
\end{table}

The network was trained to minimize the squared loss between the inferred steering angle and the ground truth (with correction from the label augmentation).
We added a L2 regularization term to avoid overfitting.
  \[
  L(Y, \hat{Y}, \theta) = L_{MSE}(Y, \hat{Y}) + \lambda_{reg} L2_{reg}(\theta) \\
  \]

\begin{table}[htbp]
\begin{center}
\caption{Network training hyperparameters}
\label{tab:hyperparams}
\begin{tabular}{c|c}
Parameter & Value \\
\hline
learning rate & $10^{-4}$, decay 0.95 by epoch \\
L2 regularization & $5.10^{-4}$ \\
Dropout & 0.8 \\
Batchsize & 100 \\
\end{tabular}
\end{center}
\end{table}

\section{Method}
This section introduces the principle of the label augmentation. Our simulator based on this label augmentation and on existing control laws is then described. Finally the different data selections on the initial dataset are detailed.

\label{sec:method}

\subsection{Label augmentation}
\label{sec:label_augmentation}
The goal of label augmentation is to simulate images that would be produced if the car was slightly translated and rotated from the original position, and add them to the training data. 
From the original recorded images, an artificial image corresponding to a slight displacement is generated from the fisheye camera (see \ref{sec:fisheye}), then the corrected label is computed using classical control laws (see \ref{sec:label-correction})


\subsubsection{Using fisheye data}
\label{sec:fisheye}
The raw fisheye image is not directly used to predict the steering wheel angle because of the distortions induced by the very large field of view. Instead, it is projected into a cylinder perpendicular to the ground plane. This projection is invariant to the camera orientation in the car and so we get homogeneous images from our 3 cars, given that the camera height is similar. Finally, the image was cropped and resized to a lower resolution using an inter area interpolation, see samples with different resolution in Figure~\ref{fig:different_size}.

With this cylindrical projection, simulating car rotations around the Z axis is trivial: the image is shifted laterally, and the lateral shift is proportional to the rotation angle.

\begin{figure}[htpb]
\centering
\begin{tabular}{ll}
\subfloat[Highway, original]{
\includegraphics[width=0.4\columnwidth]{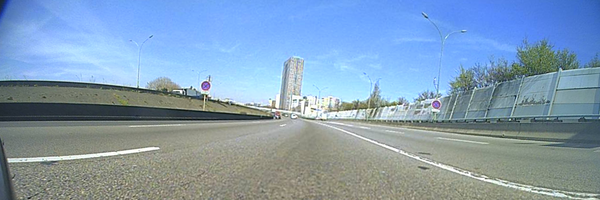}} & \subfloat[Urban, original]{
\includegraphics[width=0.4\columnwidth]{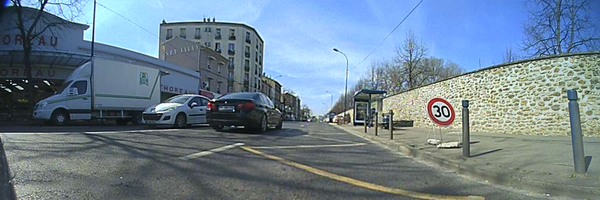}} \\
\subfloat[Highway, +1m simulated]{
\includegraphics[width=0.4\columnwidth]{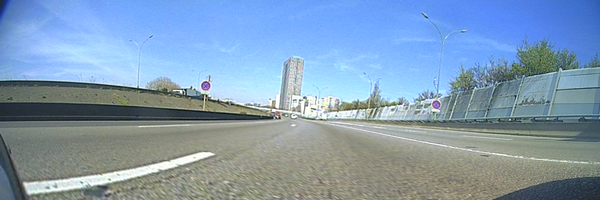} \label{fig:left-road}} & \subfloat[Urban, +1m simulated]{
\includegraphics[width=0.4\columnwidth]{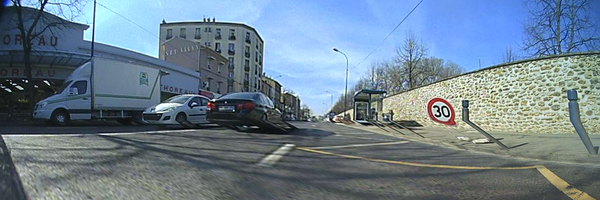} \label{fig:left-car}} \\
\subfloat[Highway, -1m simulated]{
\includegraphics[width=0.4\columnwidth]{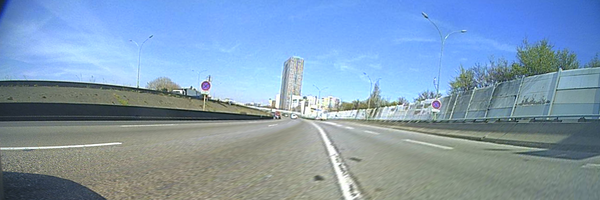} \label{fig:right-road}} & \subfloat[Urban, -1m simulated]{
\includegraphics[width=0.4\columnwidth]{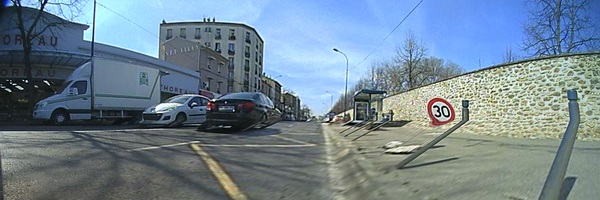} \label{fig:right-car}}
\end{tabular}
\caption{Example of different lateral offsets augmentation (highway and urban environments)}
\label{fig:translation_example}
\end{figure}

Simulating a lateral translation requires additional hypotheses. In practice, since the camera calibration is known, only one world coordinate is enough to simulate a translation, because it removes the ambiguity of projecting from image space to metric space. Here, we make the assumption that all points below the horizon are on the ground (z=0) and all points over it are at an infinite distance. This way, an image from a camera shifted in the Y direction can be generated by computing the world X and Y position of every point below the horizon, subtracting the shift from the Y value and taking the corresponding pixel from the source image.

This generates some artifacts when this assumption is not respected, for example on other road users, but not on the section relevant for the control, which is the road and the far perspective. Figure~\ref{fig:translation_example} illustrates the impact of different lateral offsets on the projection: on \ref{fig:left-road} and \ref{fig:right-road}, the deformations are barely visible, on \ref{fig:left-car} and \ref{fig:right-car} they can be seen under the horizon on vertical objects. Globally, the appearance is more realistic than the shear proposed in \cite{Hubschneider2017b}. We argue that these deformations have a limited impact on the training compared to the improvement of doing label augmentation.

\subsubsection{Label augmentation range}
The more the virtual point of view is moved away from the real one, the more the deformation induced by the image generation process is visible. Additionally, on some frames (particularly in multi-lanes condition) when the translation chosen is too large, the car seems to be on another lane, so the corrected label we are giving is not consistent. Finally, in practice, when the network is driving in inference, the car never encounters a deviation of more than 5 degrees at wheel angle compared to the human trajectory: indeed, at most common speeds, the car goes off-road before the angular deviation reaches such value. This is why  label augmentation has been performed within a limited range to avoid encountering these issues: zero-centered Gaussian distributions, with a mean and standard deviation of 0.45m and 5 degrees for translation and rotation respectively.

\subsubsection{Corrected label}
\label{sec:label-correction}
To generate the labels corresponding to the generated images, a simple lateral controller, inspired by \cite{thrun2007} \cite{Snider2009}, is used to model the human driving. It is assumed that the driver is following a trajectory (the center of the lane) using the following action on the steering wheel:
\begin{align*}
    \delta_{h}(t) &= f(\kappa(t),v) + K_e(v) e(t) + K_\theta(v) \theta(t)
\end{align*}
where $\delta_{h}(t)$ is the steering wheel angle of the human driver, $f(\kappa(t),v)$ is a function of the road curvature $\kappa(t)$ and the vehicle speed $v$, $e(t)$ and $\theta(t)$ are respectively the lateral position and angular errors that the driver is trying to minimize. Finally, $K_e(v)$ and $K_\theta(v)$ are control gains in lateral error and angular error respectively. Their values are set using existing lateral controllers developed for automated vehicles. In the case of the target test cars, $K_e(v) =  12/v \  m^{-1}$ and $K_\theta(v) =  5.3$.



The augmented data is then generated with a different error in position $e(t) + \Delta e$ and angle $\theta(t) + \Delta \theta$ with regard to the recorded data. The differences in position $\Delta e$ and angle $\Delta \theta$ are used to generate fisheye data and the corresponding label $\delta_{a}$ (steering angle).
The new label $\delta_{a}$ is simply obtained by adding $K_e(v) \Delta e  + K_\theta(v) \Delta \theta$ to the steering angle $\delta_{h}(t)$ of the  non-augmented data. This makes it possible to generate $\delta_a(t)$ directly from $\delta_h(t)$ without calculating the unknown data $f(\kappa(t), v)$, $e(t)$, $\theta(t)$.


\subsection{Our simulator}

We argue that label augmentation is the key to make imitation learning viable and that the mean error (squared or absolute) is not a valid metric to compare performance because this metric does not take into account error accumulation. A network using label augmentation could be outperformed (with this metric) by the same network trained without label augmentation, but the second one will fail to drive a real car. As a consequence, the actual impact of the label augmentation can only be measured with the inference in the loop
for the steering control. However, there is currently no online steering control benchmark to compare with existing methods, in particular \cite{Bojarski2016d}. To compensate for that, we provide quantitative performance
measurements in our simulator and qualitative estimation with one of our test cars in the loop. In practice, our simulator, though simple, is accurate enough to get a realistic estimation of the network-in-the-loop behavior of the real car.

One ideal metric would be to let the network drive a real car on diverse scenarios and record the time it can drive without any human recovery, but this is dangerous to do in practice. This is why we have designed a simulator based on real images, inspired by the idea from Nvidia \cite{Bojarski2016d} (and to our knowledge no other article used this tool). This simulation is based on the dataset and the label augmentation principle presented earlier. Using this simulator, we can obtain quantitative performance assessment of each network we trained, see \ref{sec:validation_set} for more details. The metric we used is the 'percentage of autonomy'  introduced by Nvidia \cite{Bojarski2016d}.

\begin{figure}[h]
\centering
\begin{center}
\begin{tikzpicture}[node distance=1.0cm]
\node[input, align=center] (seqim) {Current \\ image};
\node[process, right of=seqim, anchor=west, align=center,xshift=2.5cm] (def) {Image \\ transformation};
\node[process, right of=def, anchor=west, xshift=0.8cm] (cnn) {Trained \\ network};
\node[calc, below of=cnn, anchor=north, align=center, yshift=-0.35cm] (nbicycle) {Network \\ bicycle model};
\node[input, below of=seqim, align=center, anchor=north, yshift=-0.5cm](seqang) {Current \\ recorded angle};
\node[calc, right of=seqang, anchor=west, align=center,xshift=0.25cm] (hbicycle) {Human \\ bicycle model};
\node[circle, right of=hbicycle, anchor=west, xshift=0.7cm, draw=black, font=\huge] (diff) {\textbf{-}};
\draw[arrow] (seqim) -- (def);
\draw[arrow] (def) -- node[anchor=south, align=center] {Simulated \\ point \\ of view} (cnn);
\draw[arrow] (cnn) -- node[anchor=west, align=center] {Predicted \\ angle} (nbicycle);
\draw[arrow] (nbicycle) -- node[anchor=north, align=center] {Network \\ position} (diff);
\draw[arrow] (diff) -- node[anchor=west, align=center] {Offset between \\ human and network \\ positions $\Delta e \Delta \theta$} (def);
\draw[arrow] (seqang) -- (hbicycle);
\draw[arrow] (hbicycle) -- node[anchor=north, align=center] {Human \\ position} (diff);
\end{tikzpicture}
\end{center}
\caption{Simulator architecture: recorded images are transformed by the difference between recorded trajectory
and network-in-the-loop trajectory}
\label{fig:simulator_architecture}
\end{figure}
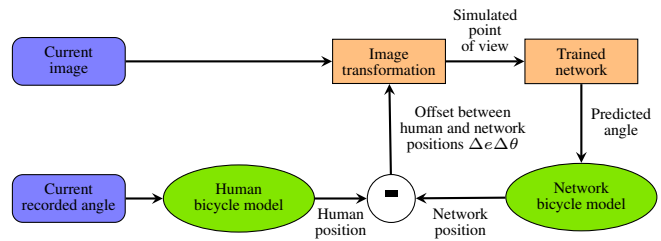

The simulator itself is based on the bicycle model \cite{Borrelli2015,Snider2009} which is commonly used for lateral control application due to its simplicity and its ability to represent correctly the vehicle dynamics. 
Here, the bicycle model is used to evaluate the behavior of the network compared to the recorded data. To this end, two bicycle models are running in parallel: one which estimates the position of the car induced by the recorded human behavior, and one with the steering prediction network in the loop. The first one is estimating the dynamics and position of the true vehicle from the actual steering angle and velocity of the human driver. The second bicycle model computes how the vehicle would behave if the network was actually controlling it. For that purpose, the input image is modified according to the difference of position between the bicycle models using the same process as the label augmentation technique described in \ref{sec:label_augmentation}. This modified image is given to the network in inference mode, which produces the corresponding steering angle. This angle is then fed to the second bicycle to get the new position of the virtual car, and so on until the end of the sequence is reached or the virtual car deviates too much and a recovery is required (see section~\ref{sec:validation_set} for the exact definition).
This is illustrated in Figure~\ref{fig:simulator_architecture}.

\begin{figure}[h]
\centering
\includegraphics[width=0.95\columnwidth]{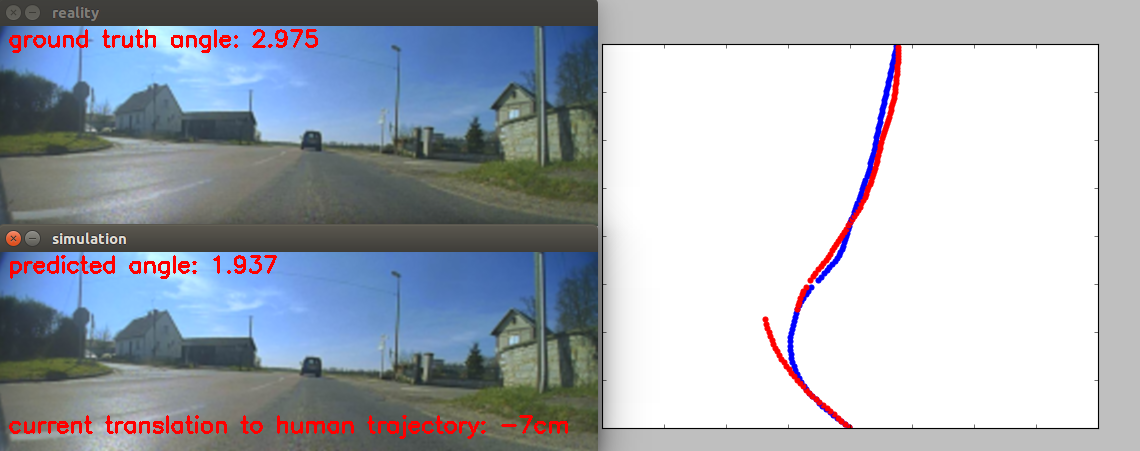}
\caption{Screenshot of the simulator, blue trajectory is human, red is the car driven by the network. At current time, the network is translated of 7cm on the right. Here the network failed to take the turn, the red trajectory goes off-road, and a recovery is done.} 
\label{fig:simulator}
\end{figure}

Figure~\ref{fig:simulator} is a screenshot of the simulator running: the actual and predicted trajectories are displayed, with the generated image from the current trajectory deviation.

\subsection{Data selection}
In the original dataset, more than 90\% of the steering angles are within the segment [-10, 10] degree (which correspond to around [-0.5, 0.5] degree for actual wheel angle).
This is why a network trained on the full dataset is biased toward going straight. To avoid this bias, we evaluate three different data selection on the training set. On ``Selection 1", around 50\% of the straight angles are discarded from the dataset. On ``Selection 2", around 85\% of the straight angles are removed. The goal is to see how much filtering is needed to get a correct performance both in turns and straight roads. Finally, a third ``oversampled"  distribution is tested where images with high steering angle are oversampled, in order to test if it is improving performance on turns. Those distributions can be characterized with their standard deviation (std), see Table~\ref{fig:steering_distribution}.


\begin{table}[h]
\centering
\caption{Characteristics of the data selections: standard deviation (std) of the steering angles \& number of samples in the [-5\degree, 5\degree] range}
\label{fig:steering_distribution}
\begin{tabular}{|c|c|c|}
\hline
Selection & Steering std & Small angle count \\
\hline
Original & 21\degree & 5M \\
Selection 1 & 26.4 \degree & 2.6M \\
Selection 2 & 35.3 \degree & 0.9M \\
Oversampled & 56 \degree & 0.9M \\
\hline
\end{tabular}
\end{table}

The impact of different field of view and crop on the input image is also evaluated, to determine the best projection to be used.

All the results of these different choices will be discussed in \ref{sec:result}.

\subsection{Fine tuning on test track}
To finetune, the previous weights with the best generalization capacity are used as an initialization, and the learning is done with a smaller initial learning rate (10 times lower than the initial one). The goal is to avoid having to record a large dataset on these use cases, while ensuring demonstration robustness. Some data from the initial dataset is kept to ensure some generalization capacity. It is interesting to note that the
finetuning recordings were not performed on the same day or time than the actual tests, and that the network still generalized and performed without failure.


\section{Results}
We first evaluate our models quantitatively using our simulator. We will describe the metric used and we present the quantitative results of different models. We then show more qualitative result on unseen environment, first on a new simulation, then on real car on open road either in France or in USA.
\label{sec:result}

\subsection{Building the test sequences for the simulator}
\label{sec:validation_set}

\begin{table}[h]
\begin{center}
\caption{Description of test scenarios}
\label{tab:valid-data}
\begin{tabular}{|c|c|c|c|}
\hline
Scenario & Urban & Highways & Sharp turns \\
\hline
Image count & 100000 & 70000 & 15000 \\
\hline
Duration (min) & 56 & 39 & 8 \\
\hline
\end{tabular}
\end{center}
\end{table}

To compare our different tests during the development and to have an idea about the behavior of the car driven by the network, we built a representative test dataset.
For this purpose, sequences from the initial test set are manually selected to represent different conditions.
In the end, the test set contains three main testing scenarios; urban, not urban (country roads and highways) and sharp turns in any situations (steering angle $\geq$ 100 degrees). These scenarios are described in Table~\ref{tab:valid-data}.

To quantitatively compare different algorithms, all of them are tested on the simulator. First, the number of recoveries in each scenario is computed. When the car driven by the network is too far from the human trajectory, we assume a recovery is needed and put it back on the human driver position: we chose one meter to be the maximum allowed distance with the human trajectory as in \cite{Bojarski2016d}. We can then compute the percentage of autonomy using the metric introduced by Nvidia \cite{Bojarski2016d} in each scenario and for each networks trained. This autonomy is defined in the following equation where $R$ is the number of recoveries, $t_r$ is the recovery time (we take 6s as in \cite{Bojarski2016d}) and $T$ is the total driving time.

\begin{equation*}
a = 1 - \frac{R\ t_r}{T}
\label{eq:autonomy}
\end{equation*}

The mean absolute distance in translation between the human and the network trajectory is also presented. But this mean distance is biased because the car back is reset to the perfect position after each recovery. In practice the autonomy was the main criterion to compare different algorithms, and the mean distance as a tie-breaker when the numbers of human recoveries were close.

\subsection{Results of different data selection}
In Table~\ref{tab:diff_distrib}, networks trained with different data selection are compared, with a baseline which is just going straight. Note that because our approach is the only one using only one fisheye camera, other
approaches from the state of the art cannot be applied here, because we cannot replicate their label augmentation techniques.
We can see from this table that the original distribution gets good results on urban and highway scenario but like expected the performance on sharp turns is not satisfactory.
Moreover removing a proportion of straight angle (selection 1 and 2) improves performance on sharp turns while keeping similar results on urban and highway (even though highway seems to be slightly impacted).

\definecolor{green}{rgb}{0,0.5,0}
\begin{table}[htpb]
\begin{center}
\caption{Autonomy (\%) and mean absolute distance (MAD, in cm) according to data distribution and validation scenario, the baseline is just going straight.}
\label{tab:diff_distrib}
\begin{tabular}{l|p{0.5cm}p{0.7cm}|p{0.5cm}p{0.7cm}|p{0.5cm}p{0.5cm}|}
Scenario & \multicolumn{2}{p{0.5cm}|}{Urban} & \multicolumn{2}{c|}{Highways} & \multicolumn{2}{c|}{Sharp turns} \\
\hline
Metric & Aut. (\%) & MAD (cm) & Aut. (\%) & MAD (cm) & Aut. (\%) & MAD (cm) \\
\hline
Original & 99.3 & 16 & \textcolor{green}{\textbf{98.7}} & \textcolor{green}{\textbf{19}} & \textcolor{red}{\textbf{73.7}} & \textcolor{red}{\textbf{30}} \\
Sel. \#1 & 98.9 & \textcolor{green}{\textbf{15}} & 97.7 & 25 & 83.7 & \textcolor{green}{\textbf{27}} \\
Sel. \#2 & \textcolor{green}{\textbf{99.5}} & 16 & 97.2 & 24 & \textcolor{green}{\textbf{87.5}} & 28 \\

Oversamp. & \textcolor{red}{\textbf{98}} & \textcolor{red}{\textbf{18}} & \textcolor{red}{\textbf{91.8}} & \textcolor{red}{\textbf{29}} & 82.5 & 29 \\

\hline
Baseline & 8 & 36 & 14 & 41 & 0 & 35 \\
\end{tabular}
\end{center}
\end{table}

On the other side, the oversampled distribution yields worse results than all others on highway and urban. Even on sharp turns, it is even worse result than both selection 1 and selection 2. We assume that the oversampled distribution goes too far, i.e. the oversampled distribution diverges too much from the original distribution (see Table \ref{fig:steering_distribution}).



\begin{figure}[h]
\centering
\subfloat[90 deg field of view]{
\includegraphics[width=0.386\columnwidth]{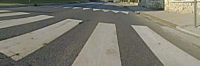}}
\subfloat[135 deg field of view, less cropping]{
\includegraphics[width=0.584\columnwidth]{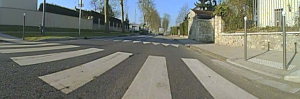}}
\caption{Samples of the image sizes tested on the same scene}
\label{fig:different_size}
\end{figure}

Testing on different fields of view and crops (Figure \ref{fig:different_size}) showed that a bigger field of view leads to a higher count of recoveries, around 80\% more in each situation. We assume that even if the network could theoretically learn to ignore some part of the image, it can lead to better learning to provide only useful information, i.e only the road and front of the camera. This experiment confirms our initial choice to crop the sky. 

\subsection{Bagging of different networks}

\begin{table}[h!]
\begin{center}
\caption{Comparison of performance between individual networks and bagging}
\label{tab:bagging}
\begin{tabular}{l|p{0.5cm}p{0.7cm}|p{0.5cm}p{0.7cm}|p{0.5cm}p{0.7cm}|}
Scenario & \multicolumn{2}{c|}{Urban} & \multicolumn{2}{c|}{Highways} & \multicolumn{2}{c|}{Sharp turn} \\
\hline
Metric & Aut. (\%) & MAD (cm) & Aut. (\%) & MAD (cm) & Aut. (\%) & MAD (cm) \\
\hline
Weights \#1 & \textcolor{green}{\textbf{99.5}} & 16 & 97.2 & 24 & \textcolor{green}{\textbf{87.5}} & 28 \\
Weights \#2 & 98.9 & 15 & 97.7 & 25 & 83.7 & \textcolor{green}{\textbf{27}} \\
Weights \#3 & 99.3 & 16 & \textcolor{green}{\textbf{98.7}} & \textcolor{green}{\textbf{19}} & \textcolor{red}{\textbf{73.7}} & \textcolor{red}{\textbf{30}} \\
Weights \#4  & 98.6 & \textcolor{red}{\textbf{18}} & \textcolor{red}{\textbf{92}} & \textcolor{red}{\textbf{26}} & 85 & 29 \\
Weights \#5  & \textcolor{red}{\textbf{98.4}} & 15 & 96.4 & 21 & 83.7 & 28 \\
\hline
Bagging  & \textcolor{green}{\textbf{99.5}} & \textcolor{green}{\textbf{13}} & \textcolor{green}{\textbf{98.7}} & \textcolor{green}{\textbf{19}} & \textcolor{green}{\textbf{87.5}} & \textcolor{green}{\textbf{27}}\\
\end{tabular}
\end{center}
\end{table}

Averaging multiple algorithms trained from different subsets of the training set can lead to an improvement in performance, this is an ensemble learning method called bagging. This improvement can be observed in the simulator. However the boost is larger when taking the average of networks trained with different parameters for data selection, which differs from the standard bagging because the different subsets are drawn from different distributions on the initial dataset. Table~\ref{tab:bagging} shows an example of such results. All individual weights are trained on a different training set: different speed selection and steering angle distributions. We can see that on all validation scenarios, the performance is better for the bagging than any individual weights.

\subsection{Tests on Grand Theft Auto (GTA) as simulator}

One interesting intermediary result was to test it in the video game GTA as simulator.
The network was able to drive in GTA when it was trained only on real image and never seen an image of this world before. The video is available at \url{https://youtu.be/y8yJQ0jGnco}.
This strong result proves the generalization capacity of the network and led us to real car testing.

\subsection{Open road results on real car}

Once a satisfactory performance was reached on the simulator, the network was integrated in a real car, to demonstrate that it was able to drive robustly on open road. However, for the first tests on an active car, the results were poorer than expected. This was due to the camera calibration of the test car which was different from the one of the training car. Indeed, we noticed that the network was very sensitive to calibration variation. To handle this issue, the simulator was used as an automatic way to select the best calibration parameters (the idea was to obtain similar images for both train and test car), which proved efficient.

We then did open road testing on places never seen in the training set (video available here \url{https://youtu.be/arBrxGDXBxQ}, the speed is controlled by the human driver).
We noticed that the network was robust to different light and weather conditions.
We also tested our network successfully on another test car in USA, even if the training set was only recorded in France.

\subsection{Fine tuning on test track on real car}
To test the impact of the fine tuning on the generalization, the performance of the finetuned models is evaluated on our simulator. Even with a high proportion of images coming from the test track (90\%), the finetuned network still has relevant performance on the general validation: about twice more recoveries than the regular one, but still 20 times less than the straight baseline.

The fine tuning is more sensitive to light condition than the generalization network so finetuning data was recorded at different times of the day and under different weather conditions to ensure a robust behavior.
We can explain this huge sensibility to light condition by the fact that the initial dataset contains lot of different light or weather conditions.
Finally the finetuned network is capable of running more than 50 laps without any human recovery and under various lightning and weather conditions. A video demonstrating the final performance of the car in CES 2018 at Las Vegas is attached to this article or available online \url{https://youtu.be/wqXR71qVZk4}. Note that for this demonstration, the speed of the car is controlled by another imitation learning network.


\subsection{Visualization of the features detected by the network}

\begin{figure}[htpb]
\centering
\subfloat[With markings]{
\includegraphics[width=0.45\columnwidth]{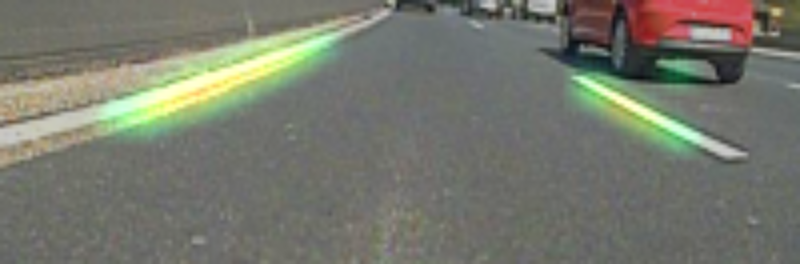}}
\subfloat[Without markings]{
\includegraphics[width=0.45\columnwidth]{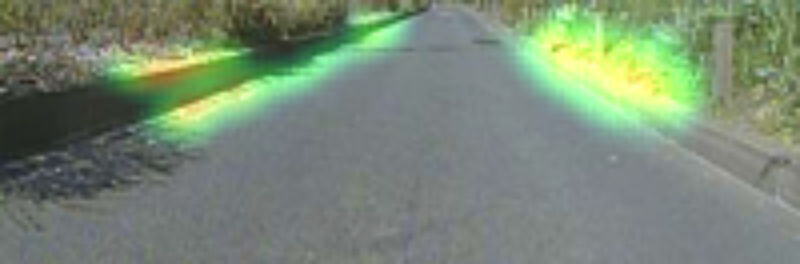}} \\
\subfloat[With traffic cones]{
\includegraphics[width=0.45\columnwidth]{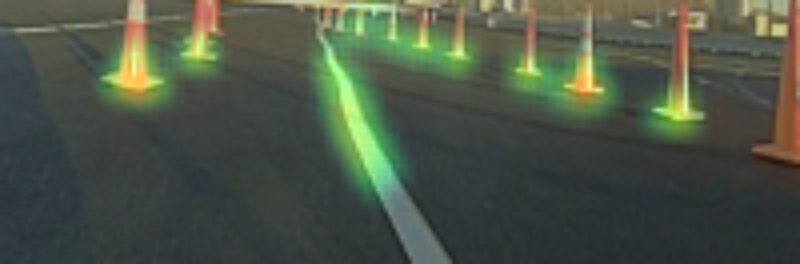}}
\caption{Samples of visualization on input images}
\label{fig:visual_backprop}
\end{figure}

Visual backpropagation \cite{2016VisualBackprop} is useful to have a better idea qualitatively of how the system is taking steering decisions. Figure \ref{fig:visual_backprop} shows some examples of this visualization. We can see that the network learned to detect road markers, borders of the road and even working cones (for the finetuned network) with only the steering angle as training signal.

\section{Conclusion}
We have presented an end-to-end network which takes as input a fisheye camera image to predict the steering angle. Taking advantage of the large field of view, a specific label augmentation procedure is designed, relying on camera model to generate realistic images for lateral control failure cases. These images are labeled using a lateral control law to get an accurate correction label. This way, failure cases are generated more robustly than what is currently used in the state of the art.

This generation is combined with a vehicle model to build a realistic simulator using real-world videos, which is then used to predict the car behavior with the control in the loop. This is opposed to pure offline approaches where the feedback from the prediction is not integrated. We use our simulator to estimate the percentage of autonomy based on frequency of recovery required by a large deviation from the desired trajectory. 
Using this metric, we define relevant data selection, label augmentation and bagging process. We also show in our network-in-the-loop realistic simulator that our trained end-to-end CNN is capable of more than 99\% autonomy on urban roads. We also validate this qualitatively on a real car, both on open road and on a test track with difficult use cases such as sharp turns or working zone areas.

In future work, better performance could be achieved by improving the neural network architecture (particularly using batch normalization \cite{2015Ioffe} and skip connections \cite{2016He}). Another interesting extension would be to handle obstacle avoidance, either by investigating in mediated perception techniques, using conditional networks \cite{Codevilla2018} or adding a temporal aspect to the prediction. 

\section*{ACKNOWLEDGMENT}
The authors would like to thank: Etienne Perot for launching this whole project, the Valeo US team for the support, Thunderhills, Mortefontaine and Amiens test track staff for their support in building the different tracks.

\bibliographystyle{IEEEtran}
\bibliography{Paper_IROS}

\begin{thebibliography}{10}
\providecommand{\url}[1]{#1}
\csname url@rmstyle\endcsname
\providecommand{\newblock}{\relax}
\providecommand{\bibinfo}[2]{#2}
\providecommand\BIBentrySTDinterwordspacing{\spaceskip=0pt\relax}
\providecommand\BIBentryALTinterwordstretchfactor{4}
\providecommand\BIBentryALTinterwordspacing{\spaceskip=\fontdimen2\font plus
\BIBentryALTinterwordstretchfactor\fontdimen3\font minus
  \fontdimen4\font\relax}
\providecommand\BIBforeignlanguage[2]{{%
\expandafter\ifx\csname l@#1\endcsname\relax
\typeout{** WARNING: IEEEtran.bst: No hyphenation pattern has been}%
\typeout{** loaded for the language `#1'. Using the pattern for}%
\typeout{** the default language instead.}%
\else
\language=\csname l@#1\endcsname
\fi
#2}}

\bibitem{Perot2017}
E.~Perot, M.~Jaritz, M.~Toromanoff, and R.~D. Charette, ``{End-to-End Driving
  in a Realistic Racing Game with Deep Reinforcement Learning},'' in \emph{IEEE
  Computer Vision and Pattern Recognition Workshops}, 2017.

\bibitem{Pan2017}
X.~Pan, Y.~You, Z.~Wang, and C.~Lu, ``{Virtual to Real Reinforcement Learning
  for Autonomous Driving},'' \emph{Proceedings of the British Machine Vision
  Conference (BMVC)}, 2017.

\bibitem{Goodfellow}
I.~J. Goodfellow, J.~Pouget-Abadie, M.~Mirza, \emph{et~al.}, ``{Generative
  Adversarial Nets},'' \emph{NIPS}, 2014.

\bibitem{Bojarski2016d}
M.~Bojarski, D.~D. Testa, D.~Dworakowski, \emph{et~al.}, ``{End to End Learning
  for Self-Driving Cars},'' \emph{ArXiv preprint}, 2016.

\bibitem{Pomerleau1989a}
D.~A. Pomerleau, ``{Alvinn: An autonomous land vehicle in a neural network},''
  \emph{NIPS}, 1989.

\bibitem{Xu}
H.~Xu, Y.~Gao, F.~Yu, and T.~Darrell, ``{End-to-end Learning of Driving Models
  from Large-scale Video Datasets},'' \emph{CVPR}, 2017.

\bibitem{Eraqi2017a}
H.~M. Eraqi, M.~N. Moustafa, and J.~Honer, ``{End-to-End Deep Learning for
  Steering Autonomous Vehicles Considering Temporal Dependencies},''
  \emph{NIPS}, 2017.

\bibitem{Chi}
L.~Chi and Y.~Mu, ``{Deep Steering: Learning End-to-End Driving Model from
  Spatial and Temporal Visual Cues},'' \emph{ArXiv preprint}, 2017.

\bibitem{udacity}
\BIBentryALTinterwordspacing
``Udacity challenge and dataset.'' [Online]. Available:
  \url{https://tinyurl.com/yangl4qj}
\BIBentrySTDinterwordspacing

\bibitem{Hochreiter1997}
S.~Hochreiter and J.~{Schmidhuber}, ``{Long Short-Term Memory},'' \emph{Neural
  Computation}, 1997.

\bibitem{Hubschneider2017b}
C.~Hubschneider, A.~Bauer, M.~Weber, and J.~M. Z{\"{o}}llner, ``{Adding
  Navigation to the Equation: Turning Decisions for End-to-End Vehicle
  Control},'' in \emph{IEEE International Conference on Intelligent
  Transportation}, 2017.

\bibitem{Codevilla2018}
F.~Codevilla, M.~M{\"u}ller, A.~L{\'o}pez, V.~Koltun, and A.~Dosovitskiy,
  ``End-to-end driving via conditional imitation learning,'' in
  \emph{International Conference on Robotics and Automation (ICRA)}, 2018.

\bibitem{Yang2018b}
Z.~Yang, Y.~Zhang, J.~Yu, J.~Cai, and J.~Luo, ``{End-to-end Multi-Modal
  Multi-Task Vehicle Control for Self-Driving Cars with Visual Perceptions},''
  \emph{ArXiv preprint}, 2018.

\bibitem{thrun2007}
G.~M. Hoffmann, C.~J. Tomlin, M.~Montemerlo, and S.~Thrun, ``Autonomous
  automobile trajectory tracking for off-road driving: Controller design,
  experimental validation and racing,'' in \emph{American Control Conference},
  2007.

\bibitem{Snider2009}
J.~M. Snider, ``{Automatic Steering Methods for Autonomous Automobile Path
  Tracking},'' 2009.

\bibitem{Borrelli2015}
J.~Kong, M.~Pfeiffer, G.~Schildbach, and F.~Borrelli, ``Kinematic and dynamic
  vehicle models for autonomous driving control design,'' in \emph{IEEE
  Intelligent Vehicles Symposium (IV)}, 2015.

\bibitem{2016VisualBackprop}
M.~{Bojarski}, A.~{Choromanska}, K.~{Choromanski}, \emph{et~al.},
  ``{VisualBackProp: efficient visualization of CNNs},'' \emph{ArXiv preprint},
  2016.

\bibitem{2015Ioffe}
S.~{Ioffe} and C.~{Szegedy}, ``{Batch Normalization: Accelerating Deep Network
  Training by Reducing Internal Covariate Shift},'' \emph{Proceedings of The
  32nd International Conference on Machine Learning}, 2015.

\bibitem{2016He}
K.~{He}, X.~{Zhang}, S.~{Ren}, and J.~{Sun}, ``{Deep Residual Learning for
  Image Recognition},'' \emph{CVPR}, 2016.

\end{thebibliography}

\end{document}